\documentclass{isprs} 
\usepackage{subfigure}
\usepackage{setspace}
\usepackage{geometry} 
\usepackage{epstopdf}
\usepackage[labelsep=period]{caption}  
\usepackage[british]{babel} 
\usepackage[hang]{footmisc}
\usepackage{amssymb}
\usepackage{url}
\usepackage{pifont}
\usepackage{xcolor}
\usepackage{graphicx}

\newcommand{\cmark}{\ding{51}}
\newcommand{\3}[1]{\textcolor{blue}{#1}}
\newcommand{\1}[1]{\textcolor{red}{#1}}
\newcommand{\2}[1]{\textcolor{green}{#1}}

\begin{document}

\title{Bidirectional Multi-scale Attention Networks for Semantic Segmentation of Oblique UAV Imagery}
\version{}

\author{Ye Lyu\textsuperscript{a}, George Vosselman\textsuperscript{a}, Gui-Song Xia\textsuperscript{b}, Michael Ying Yang\textsuperscript{a}}
\address{\textsuperscript{a}University of Twente, The Netherlands\\\textsuperscript{b}Wuhan University, China}


\icwg{}   

\abstract{
Semantic segmentation for aerial platforms has been one of the fundamental scene understanding task for the earth observation. Most of the semantic segmentation research focused on scenes captured in nadir view, in which objects have relatively smaller scale variation compared with scenes captured in oblique view. The huge scale variation of objects in oblique images limits the performance of deep neural networks (DNN) that process images in a single scale fashion. In order to tackle the scale variation issue, in this paper, we propose the novel bidirectional multi-scale attention networks, which fuse features from multiple scales bidirectionally for more adaptive and effective feature extraction. The experiments are conducted on the UAVid2020 dataset and have shown the effectiveness of our method. Our model achieved the state-of-the-art (SOTA) result with a mean intersection over union (mIoU) score of $70.80\%$.}


\maketitle


\section{Introduction}\label{intro}
\begin{figure}[ht!]
	\begin{center}
		\includegraphics[width=1.0\columnwidth]{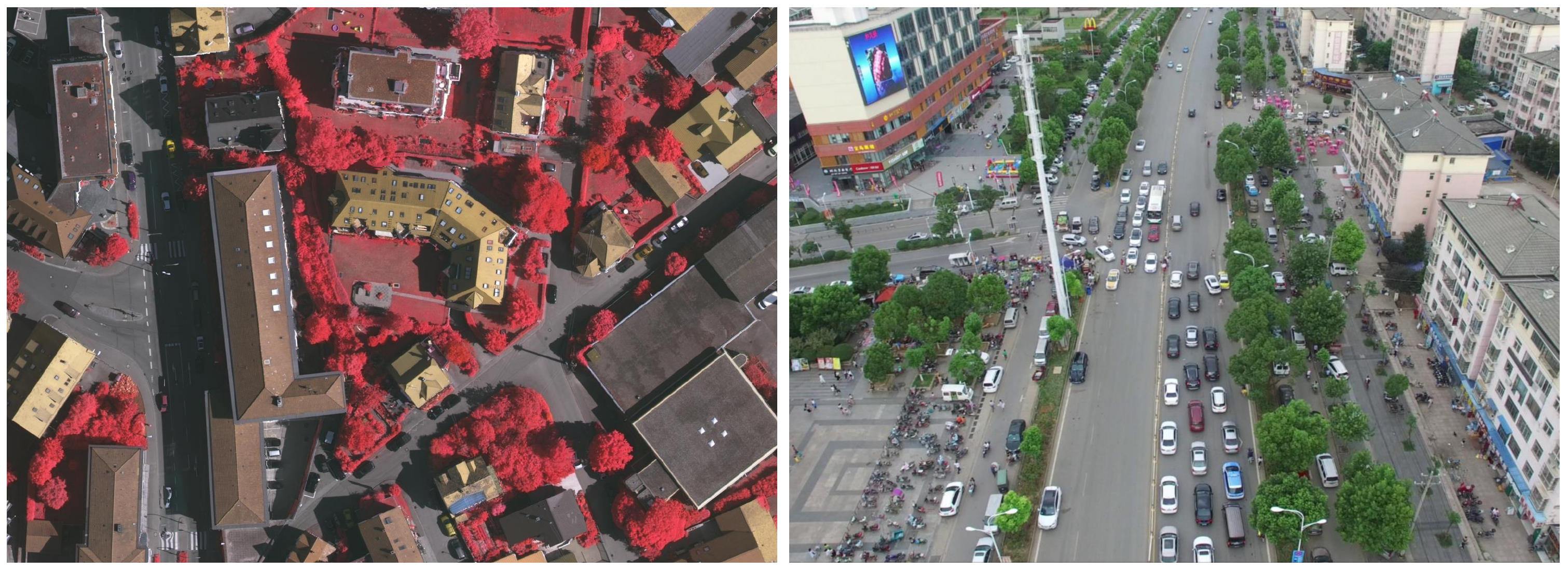}
		\caption{Example of images in different viewing style. The left image from Vaihingen dataset\protect\cite{ISPRSbenchmark} is captured in nadir view. The right image from UAVid2020 dataset\protect\cite{uavid} is captured in oblique view.}
		\label{fig:view_compare}
	\end{center}
\end{figure}

Semantic segmentation has been one of the most fundamental research tasks for scene understanding. It is to assign each pixel within an image with the class label it belongs to. There have been many works for semantic segmentation on the remote sensing images and the aerial images~\cite{deepglobe, ISPRSbenchmark}, which are captured in nadir view style. The spatial resolutions in such images are approximately the same for all pixels. Oblique views have a much larger land coverage if the platforms are at the same flight height. For example, the unmanned aerial vehicle (UAV) platform has been used to for urban scene observation~\cite{uavid,aeroscapes}. The images of different viewing directions are shown in Figure~\ref{fig:view_compare}. The left image of nadir view is from the Vaihingen dataset~\cite{ISPRSbenchmark}, while the right image of oblique view is from the UAVid2020 dataset~\cite{uavid}. Compared with the images in nadir view style, the images in oblique view have very large spatial resolution variation across the entire image.

The state-of-the-art methods for semantic segmentation all rely on powerful deep neural networks, which can effectively extract high-level semantic information to determine the class types for all pixels. Deep neural networks serve as non-linear functions, which map an image input to a label output. Due to its non-linear property, the label output will not scale linearly as the image input scales. When designing the deep neural networks, there is usually a performance trade-off for objects in different scales. For example, the semantic segmentation of a small car in a remote sensing image is better handled in higher resolution where finer details can be observed, such as wheels. For larger objects like roads and buildings, it is better to have more global context to recognize the objects since their whole shapes can be observed for semantic segmentation.

When objects in an image dataset have very large scale variation, the semantic segmentation performance of deep neural networks will drop if this multi-scale problem is not considered in the network design. A simple strategy is to apply multi-scale inference~\cite{pspnet}, i.e., a well-trained deep neural networks predict the score maps of the same image in multiple different scales, and the score maps are averaged to determine the final label prediction. Such strategy generally provides better performance. However, a good prediction from a proper scale could be undermined by those worse predictions from other scales, which limits the model performance. Max-pooling selects one score map prediction of multiple scales for each pixel, but the optimal output could be the interpolation of the prediction of multiple scales. A smarter way of fusing the output score maps is to leverage on an attention model~\cite{attn2scale}, which determines the weights when fusing the score maps of different scales. The strategy has been extended to a hierarchical structure for better performance~\cite{hmsa}.

With respect to the design of deep neural networks, there are several strategies to relieve the multi-scale problem. The first strategy is to gradually refine features from coarse scales to fine scales~\cite{fcn8s,unet,uavid}. The second strategy is to design a multi-scale feature extractor module in the middle of the deep neural networks~\cite{pspnet,deeplabv3,deeplabv3+,ocnet}.
Self-attention~\cite{danet,ccnet,ocrnet} and graph networks~\cite{graph_reason,graph_represent} have also been applied to aggregate information globally to reinforce the features for each pixel.

In this paper, we propose the bidirectional multi-scale attention networks (BiMSANet) to address the multi-scale problem in the semantic segmentation task. Our method is inspired by the multi-scale attention strategy~\cite{attn2scale,hmsa} and the feature level fusion strategy~\cite{deeplabv3,pspnet}, and jointly fuses the features guided by the attention of different scales in bidirectional pathways, i.e, coarse-to-fine and fine-to-coarse. Our method is tested on the new UAVid2020 dataset~\cite{uavid}. One of its challenges is the huge inter-class and intra-class scale variance for different objects due to its oblique viewing style. Our method achieves a new state-of-the-art result with a mIoU score of $70.8\%$. Compared with the currently top ranked method~\cite{hmsa}, which features on handling the multi-scale problem, our methods outperforms by almost $0.8\%$.

The contributions of this paper are summarized as follows,
\begin{itemize}
	\item We have proposed a novel bidirectional multi-scale attention networks (BiMSANet) to handle the multi-scale problem for the semantic segmentation task.
	\item We have visualized in multiple perspectives and analyzed the bidirectional multi-scale attentions in details.
	\item We have achieved state-of-the-art result on the UAVid2020 benchmark, and the code will be made public.
\end{itemize}

\section{Related Work}
In this section, we will discuss other works that are related to our paper. In order to handle the multi-scale problem for the semantic segmentation, a number of deep neural networks have been designed.

\textbf{Multi-scale feature fusion.}
The first basic type of method is to aggregate features of multiple scales from deep neural networks. FCN~\cite{fcn8s} and U-Net~\cite{unet} have adopted skip connections between encoder and decoder to gradually fuse the information from multiple scales. MSDNet~\cite{uavid} has extended the connection across scales to further increase the performance. ZipZagNet~\cite{zigzag} uses a more complex zip-zag architecture between the backbone and the decoder for intermediate multi-scale feature aggregation. HRNet~\cite{hrnet} proposes a multi-scale backbone to exchange information between branches of coarse scale and fine scale. BiSeNet~\cite{bisenet} proposes a dual branch structure for better performance, one branch for higher spatial resolution, while the other for richer semantic features.

\textbf{Multi-scale context extraction.}
Another method is to aggregate multi-scale context from the same feature maps with a module. PSPNet~\cite{pspnet} has adopted pyramid pooling module, which has pooling modules of multiple scales to pool context features for the object recognition. DeepLabv3~\cite{deeplabv3,deeplabv3+} has utilized atrous spatial pyramid pooling module, which assembles multi-scale features with convolutions of multiple atrous rates. OCNet~\cite{ocnet} proposes pyramid object context (Pyramid-OC) module and atrous spatial pyramid object context (ASP-OC) module to extract object context in multiple scales.

\textbf{Context by relations.}
With the creation of self-attention mechanism~\cite{attention} for natural language processing, better semantic segmentation results have also been achieved when self-attention is applied to reason the relation between pixels. Self-attention refines the features in a non-local style, which aggregates information for each pixel globally. DANet~\cite{danet} has utilized dual attention module, position attention and channel attention, to extract information globally. CCNet~\cite{ccnet} has applied the criss-cross attention module to reduce the computational complexity of the self-attention. OCRNet~\cite{ocrnet} has used explicit class attention to reinforce the features.
However, these types of methods are normally intensive in memory and computation as there are too many pixels, resulting in very dense connections between them. Graph reasoning is another way to include relations among objects. Instead of adopting dense pixel relations, sparse graph structure makes the context relation reasoning less intensive in memory and computation. ~\cite{graph_reason} proposes the symbolic graph reasoning (SGR) layer for context information aggregation through knowledge graph. ~\cite{graph_represent} transforms a 2D image into a graph structure, whose vertices are clusters of pixels. Context information is propagated across all vertices on the graph.

\textbf{Inference in multi-scale.}
Multi-scale inference is widely used to provide more robust prediction, which is orthogonal to previously discussed methods as those networks can be regarded as a trunk for multi-scale inference. Average pooling and max pooling on score maps are mostly used, but they limit the performance. ~\cite{attn2scale} propose to apply attentions for fusing score maps across multiple scales. The method is more adaptive to objects in different scales as the weights for fusing score maps across multiple scales can vary. ~\cite{hmsa} further improve the multi-scale attention method by introducing a hierarchical structure, which allows different network structures during training and testing to improve the model design.
Our paper also focuses on the multi-scale inference. We have further improved the multi-scale attention mechanism by introducing feature level bidirectional fusion.

\section{Preliminary}
In this section, we first go through some network architecture design to better help understand the newly proposed bidirectional multi-scale attention networks.
\subsection{Multi-Scale-Dilation Net}
The multi-scale-dilation net~\cite{uavid} is proposed as the first attempt to tackle the multi-scale problem for the UAVid dataset. The basic idea shares the philosophy of multi-scale image inputs, where the input images are scaled by the scale to batch operation and batch to scale operation. The intermediate features are concatenated from coarse to fine scales, which are used to output the final semantic segmentation output. The structure is shown in Figure~\ref{fig:msd}. The feature extraction part is named as trunk in the following figures.
\begin{figure}[ht!]
	\begin{center}
		\includegraphics[width=1.0\linewidth]{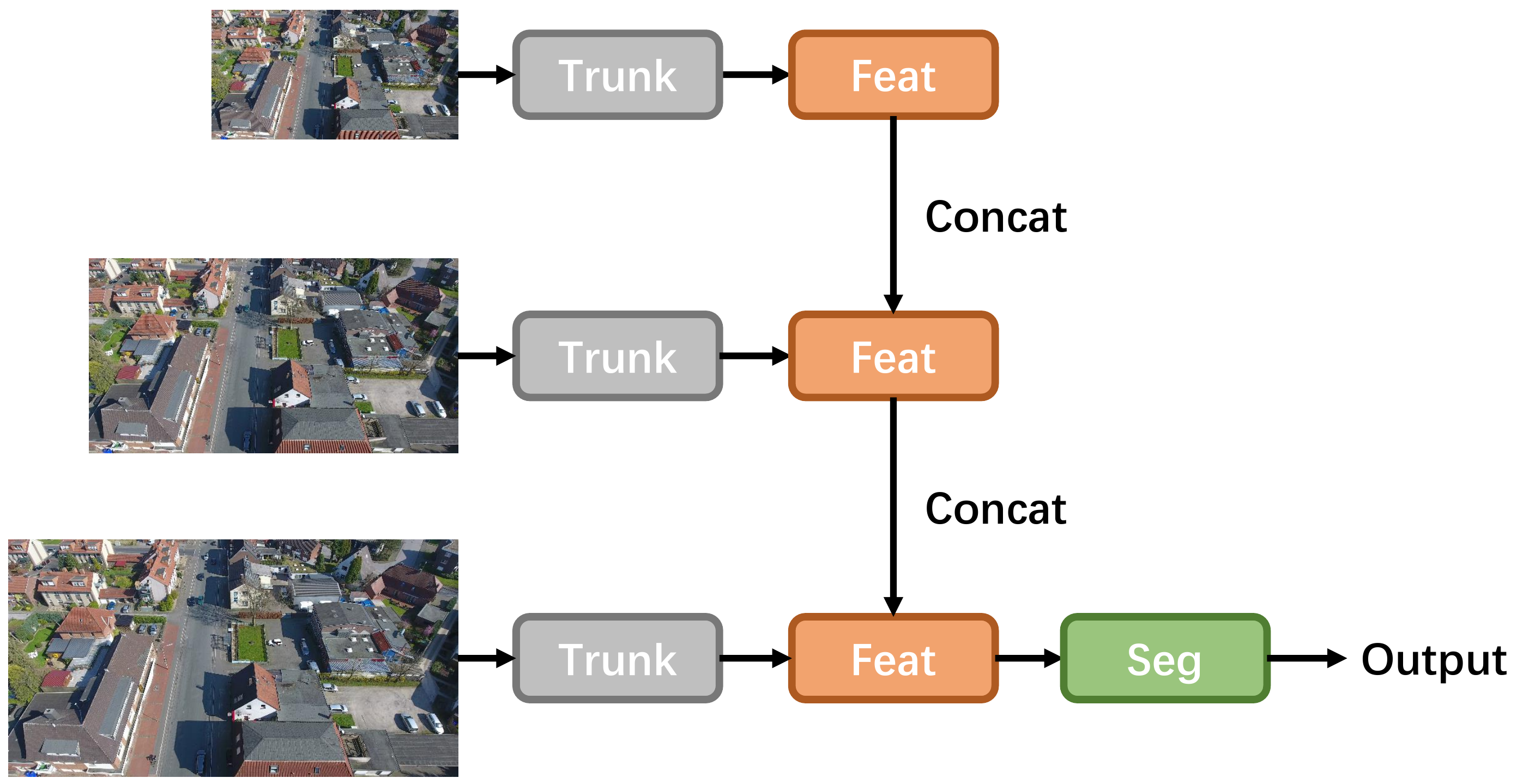}
		\caption{Architecture of the multi-scale dilation net. Features are aggregated from coarse to fine scales with concatenation.}
		\label{fig:msd}
	\end{center}
\end{figure}

\subsection{Hierarchical Multi-Scale Attention Net}
\begin{figure}[ht!]
	\begin{center}
		\includegraphics[width=1.0\linewidth]{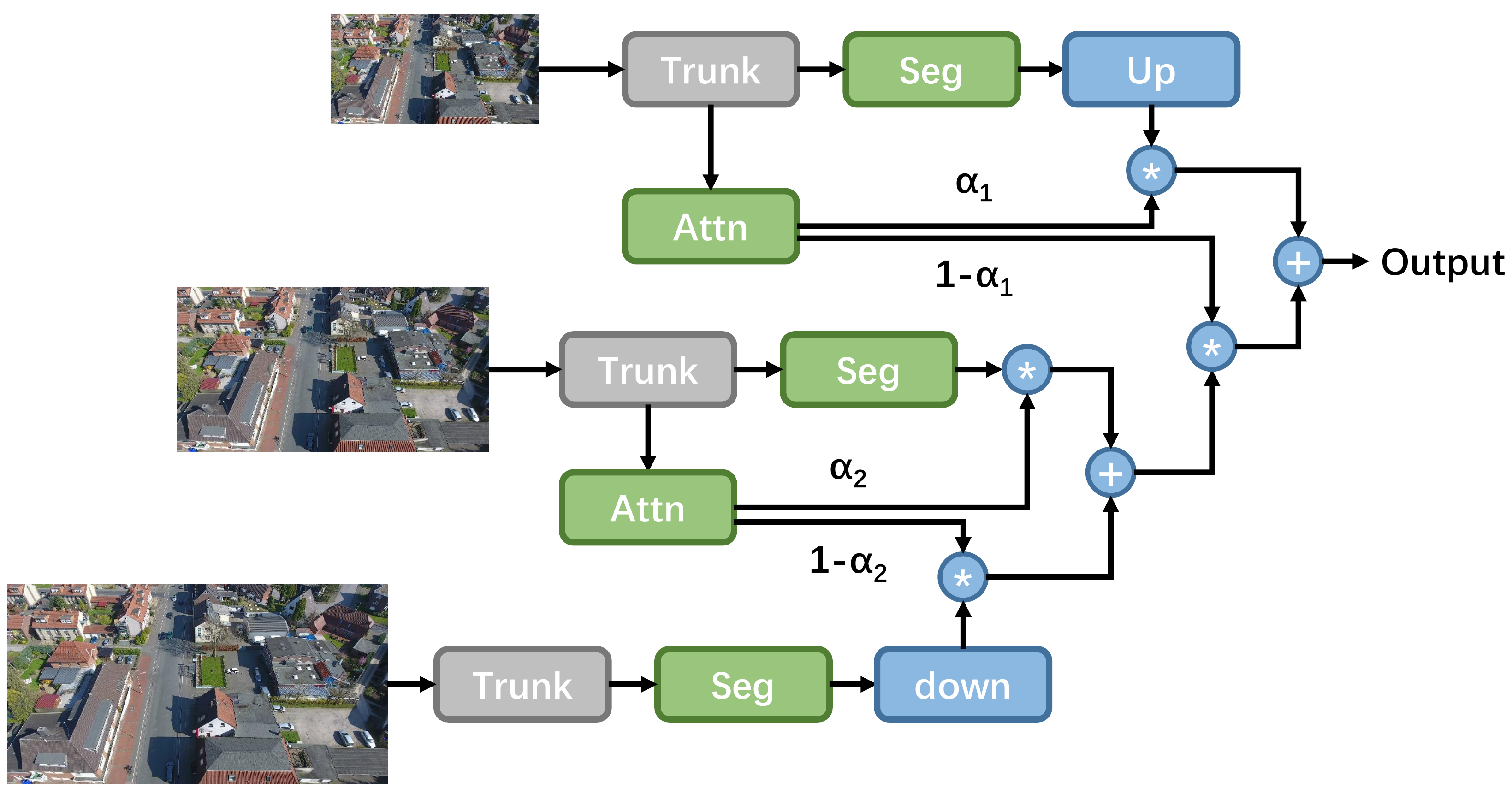}
		\caption{Architecture of the hierarchical multi-scale attention networks. In addition to the predicted score maps, extra weighting masks are predicted from the attention sub-networks for fusing the score maps of adjacent scales. $\oplus, \circledast$ stand for element-wise addition and multiplication, respectively.}
		\label{fig:hmsa}
	\end{center}
\end{figure}

The hierarchical multi-scale attention net~\cite{hmsa} is proposed to learn to fuse semantic segmentation outputs of adjacent scales by a hierarchical attention mechanism. The deep neural networks learn to segment the images while predicting the weighting masks for fusing the score maps. This method ranks as the top method in the Cityscapes pixel-level semantic labeling task~\cite{cityscapes}, which focuses on the multi-scale problem. The hierarchical mechanism allows different network structures during training and inference, e.g., the networks have only two branches of two adjacent scales during training, while the networks could have three branches of three adjacent scales during testing as shown in Figure~\ref{fig:hmsa}. 

\subsection{Feature Level Hierarchical Multi-Scale Attention Net}
\begin{figure}[ht!]
	\begin{center}
		\includegraphics[width=1.0\linewidth]{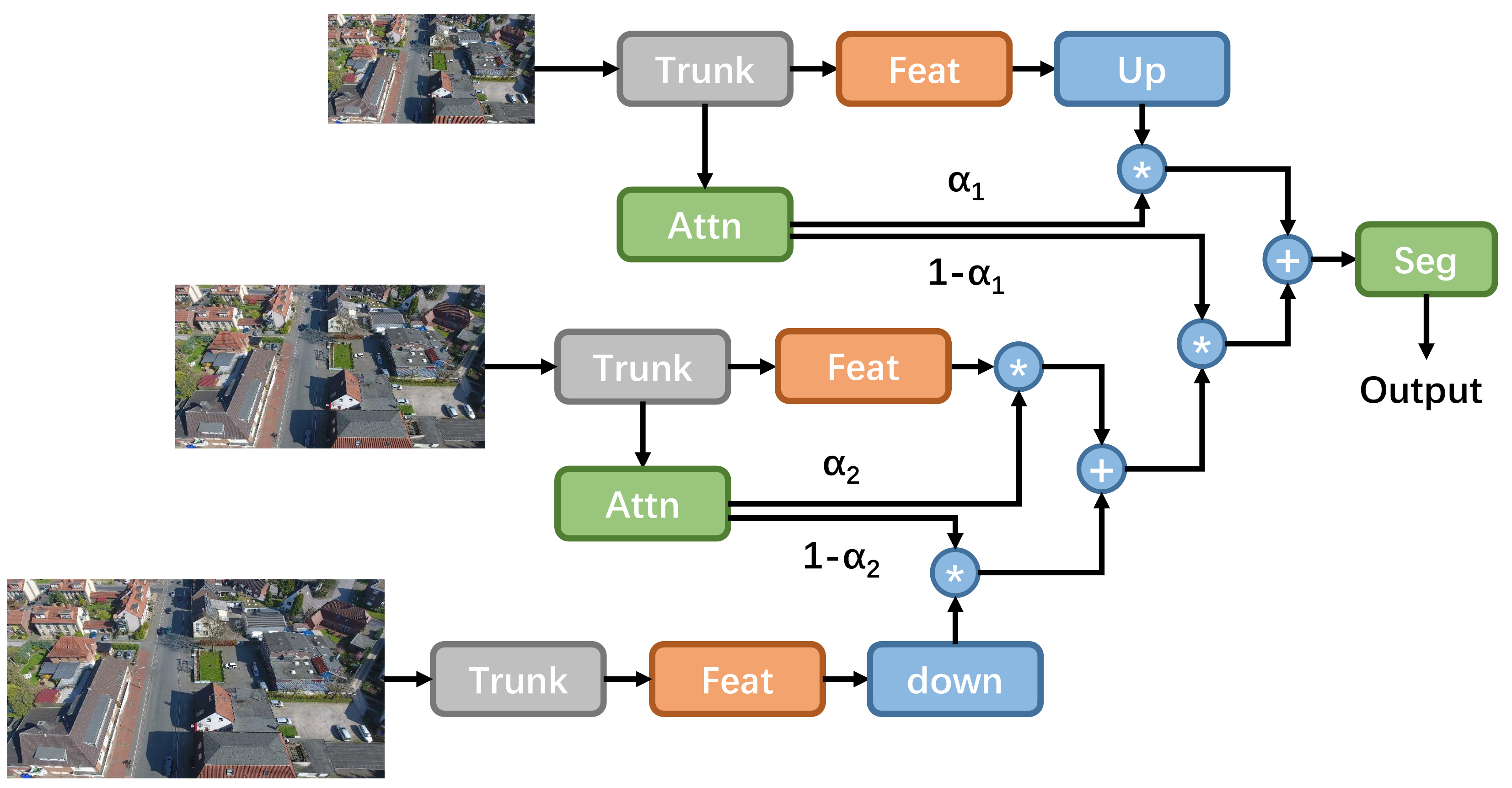}
		\caption{Architecture of the hierarchical multi-scale attention networks with feature level fusion. Segmentation head is moved to the end of the fused features. $\oplus, \circledast$ stand for element-wise addition and multiplication, respectively.}
		\label{fig:fhmsa}
	\end{center}
\end{figure}

One limitation of the hierarchical multi-scale attention networks is that the fused score maps are the linear interpolation of the score maps in adjacent scales, whereas the best score maps could be acquired with the interpolated features instead. A simple solution that we propose is to move the segmentation head to the end of the fused features as shown in Figure~\ref{fig:fhmsa}. 

\begin{figure*}[ht!]
	\begin{center}
		\includegraphics[width=0.85\linewidth]{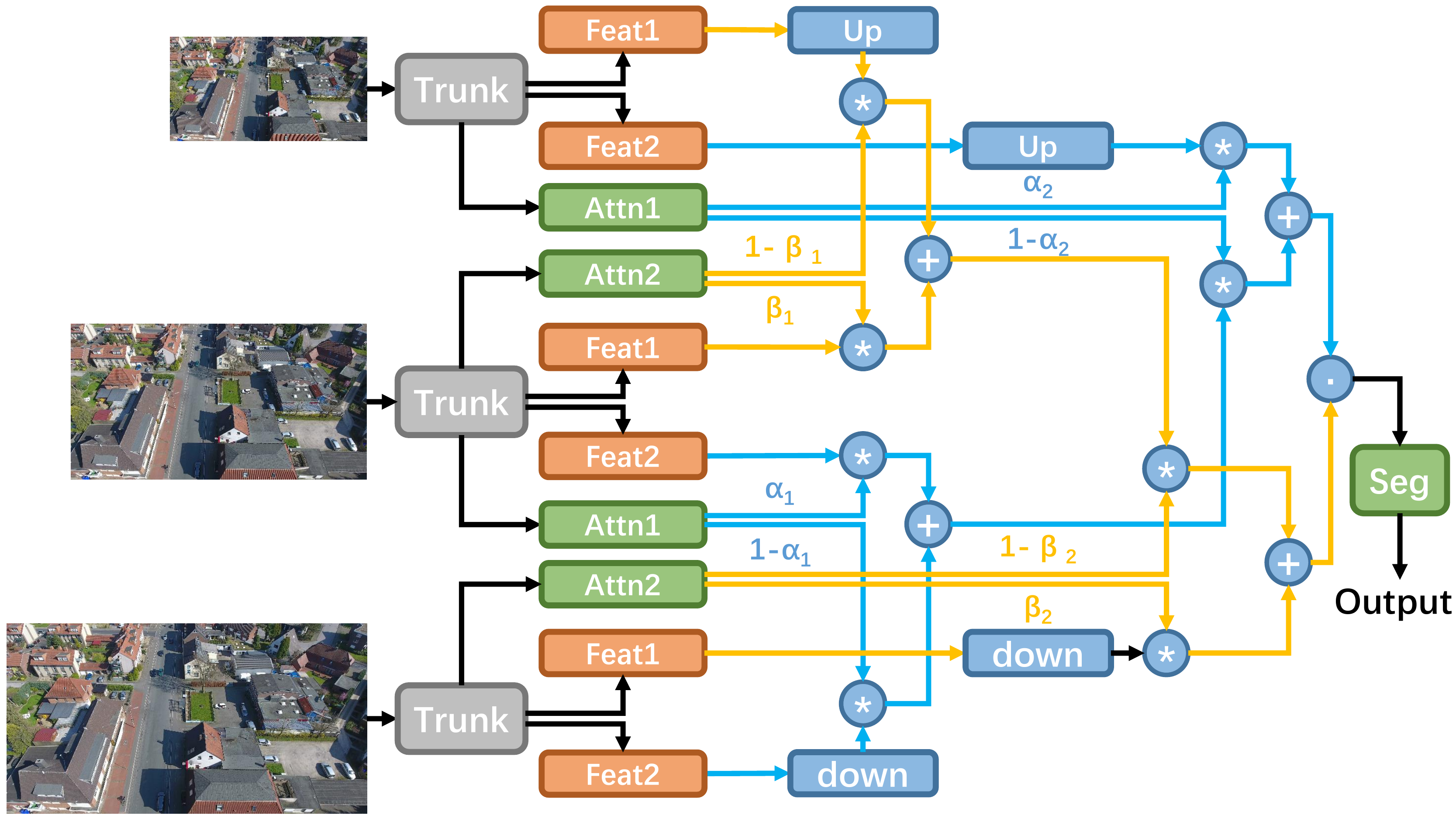}
		\caption{Architecture of the bidirectional multi-scale attention networks. The structure is the combination of two feature level hierarchical multi-scale attention nets corresponding to two pathways, where they share the same trunks. The coarse to fine pathway and the fine to coarse pathway are marked with the yellow and the blue arrows, respectively. $\oplus, \circledast$ stand for element-wise addition and multiplication, respectively. $\odot$ stands for concatenation in channel dimension.}
		\label{fig:bimsa}
	\end{center}
\end{figure*}

\section{Bidirectional Multi-scale Attention Networks}
In this section, the structure of the proposed bidirectional multi-scale attention networks will be introduced. 
\subsection{Overall Architecture}
Our design also takes the hierarchical attention mechanism and the feature level fusion into account. The overall architecture is shown in Figure~\ref{fig:bimsa}. For the input image $I$ of size $H\times W$, the image pyramid is built by adding two extra images $I_{2\times}$ and $I_{0.5\times}$, which are acquired by bi-linear up-sampling $I$ to size of $2H\times 2W$ and bi-linear down-sampling $I$ to $\frac{1}{2}H\times \frac{1}{2}W$. The bidirectional multi-scale attention networks have two pathways for feature fusing in a hierarchical manner. For each pathway, the structure is the same as the feature level hierarchical multi-scale attention nets. The design of the two pathways allows the feature fusion from both directions, and the fusion weights can be better determined in a better scale. The reason to use feature level fusion is that we need distinct features for two pathways. If the score maps are used for fusion, the feat1 and the feat2 in the two pathways would be the same, which limits the representation power of the two pathways. The two pathways take advantage of their own attention branches and features. Attn1 branch and Feat1 are for the coarse to fine pathway, while Attn2 branch and Feat2 are for the fine to coarse pathway. The Feat1 and the Feat2 from two pathways are fused hierarchically across scales, and the final feature is the concatenation of the features from the two pathways. 

The Feat1 and Feat2 are reduced to the half number of channels as the Feat in feature level hierarchical multi-scale attention net. This setting is to provide fair comparisons between these two types of networks, since it leads to features with the same number of channels before the segmentation head.

The parameter sharing is also applied in the design. Three branches corresponding to the three scales share the same network parameters for Trunk, Attn1 and Attn2. Feat1 and Feat2 in the three branches are different as they are the output of different image inputs through the same trunk. 

\subsection{Module Details}
In this section, we will illustrate the details of each component  we applies.

\textbf{Trunk.} In order to effectively extract information from each single scale, we have adopted the deeplabv3+~\cite{deeplabv3+} as the trunk. We apply the wide residual networks~\cite{wrn} as the backbone, namely the WRN-38, which has been pre-trained on the imagenet dataset~\cite{imagenet}. The ASPP module in the deeplabv3+ has convolutions with atrous rate of $1,6,12,$ and $18$. The features $f_{b}$ from the deeplabv3+ are further refined with a sequence of modules as follows, $Conv3\times3(256)->BN->ReLU->Conv3\times3(256)->BN->ReLU->Conv1\times1(nc)$ (numbers in the brackets are the numbers of output channels), which corresponds to the feature transformation in the $Seg$ of the hierarchical multi-scale attention net before the final classification.

The trunk $T$ transforms an image input $I$ into feature maps $f$ with $nc$ channels, i.e., $f=T(I)$. $nc=n_{class}\times d$, where $n_{class}$ is the total number of classes for the semantic segmentation task. $d$ is the expansion rate for the channels. $d$ is set to $4$ in our case. The first $\frac{1}{2}nc$ channels are for the Feat1, while the second $\frac{1}{2}nc$ channels are for the Feat2.

\textbf{Attention head.}
The Attn1 and the Attn2 share the same structure, but with different parameters. The attention heads map the features $f_{b}$ from the deeplabv3+ to the attention weights $\alpha,\beta$ (ranging from $0.0$ to $1.0$ with $\frac{1}{2}nc$ channels) for the two pathways. For each attention head, the structure is comprised of a sequence of modules as follows, $Conv3\times3(256)->BN->ReLU->Conv3\times3(256)->BN->ReLU->Conv1\times1(\frac{1}{2}nc)->Sigmoid$ (numbers in the brackets are the output channels).

\textbf{Segmentation head.}
The segmentation head $Seg$ converts the fused input feature maps $f_{fused}$ into score maps $l$ ($8 channels$ for the UAVid2020 dataset), which correspond to the class probabilities for all the pixels, i.e., $l=Seg(f_{fused})$.
The segmentation head is simply a $1\times 1$ convolution, $Conv1\times1(n_{class})$. Argmax operation along the channel dimension outputs the final class labels for all the pixels.

\textbf{Auxiliary semantic head.} As in ~\cite{hmsa}, we apply auxiliary semantic segmentation heads for each branch during training, which consists of only a $1\times 1$ convolution, $Conv1\times1(n_{class})$.

\subsection{Training and inference}
As our model follows the hierarchical inference mechanism, it allows our model to be trained with only $2$ scales, while to infer with $3$ scales ($0.5\times, 1\times, 2\times$). Such design makes it possible for our network to adopt a large trunk such as deeplabv3+ with WRN-38 backbone for better performance.
We use RMI loss~\cite{rmi} for the main semantic segmentation head and cross entropy loss for the auxiliary semantic head.

\section{Experiments}
In this section, we will illustrates the implementation details for the experiments and compare the performance of different models on the UAVid2020 dataset.

\subsection{Dataset and Metric}
Our experiments are conducted on the public UAVid2020 dataset\footnote{https://uavid.nl/}~\cite{uavid}. The UAVid2020 dataset focuses on the complex urban scene semantic segmentation task for $8$ classes. The images are captured in oblique views with large spatial resolution variation. There are $420$ high quality images of $4K$ resolutions ($4096\times2160$ or $3840\times2160$) in total, split into training, validation and testing sets with $200$, $70$ and $150$ images, respectively. 
The performance of different models are evaluated on the test set of the UAVid2020 benchmark.
The performance for the semantic segmentation task is assessed based on the standard mean intersection-over-union(mIoU) metric~\cite{PascalVOC}.

\begin{table*}[h]
	\centering
	\begin{tabular}{|l|c|c|c|c|c|c|c|c|c|}\hline
		Methods		&mIoU(\%)	&Clutter&Building&Road&Tree&Vegetation&Moving Car&Static Car&Human\\\hline
		MSDNet&56.97&57.04&79.82&73.98&74.44&55.86&62.89&32.07&19.69\\\hline
		DeepLabv3+&67.36&66.68&87.61&80.04&\3{79.49}&\2{62.00}&71.69&68.58&22.76\\
		HMSANet&\3{70.03}&\3{69.32}&\2{88.14}&\2{82.12}&79.42&61.21&\1{77.33}&\3{72.52}&\2{30.17}\\
		FHMSANet&\2{70.33}&\2{69.36}&\3{87.95}&\1{82.69}&\2{80.06}&\1{62.66}&\3{76.88}&\2{72.90}&\3{30.12}\\
		BiMSANet&\1{70.80}&\1{69.94}&\1{88.63}&\3{81.60}&\1{80.38}&\3{61.64}&\2{77.22}&\1{75.62}&\1{31.34}\\\hline
	\end{tabular}
	\caption{Performance comparisons in intersection-over-union (IoU) metric for different models. The top ranked scores are marked in colors. \1{Red} for the 1st place, \2{green} for the 2nd place, and \3{blue} for the 3rd place.}
	\label{tab:comparisons}
\end{table*}

\begin{figure*}[ht!]
	\centering
	\begin{minipage}{.33\linewidth}
		\includegraphics[width=1.0\linewidth]
		{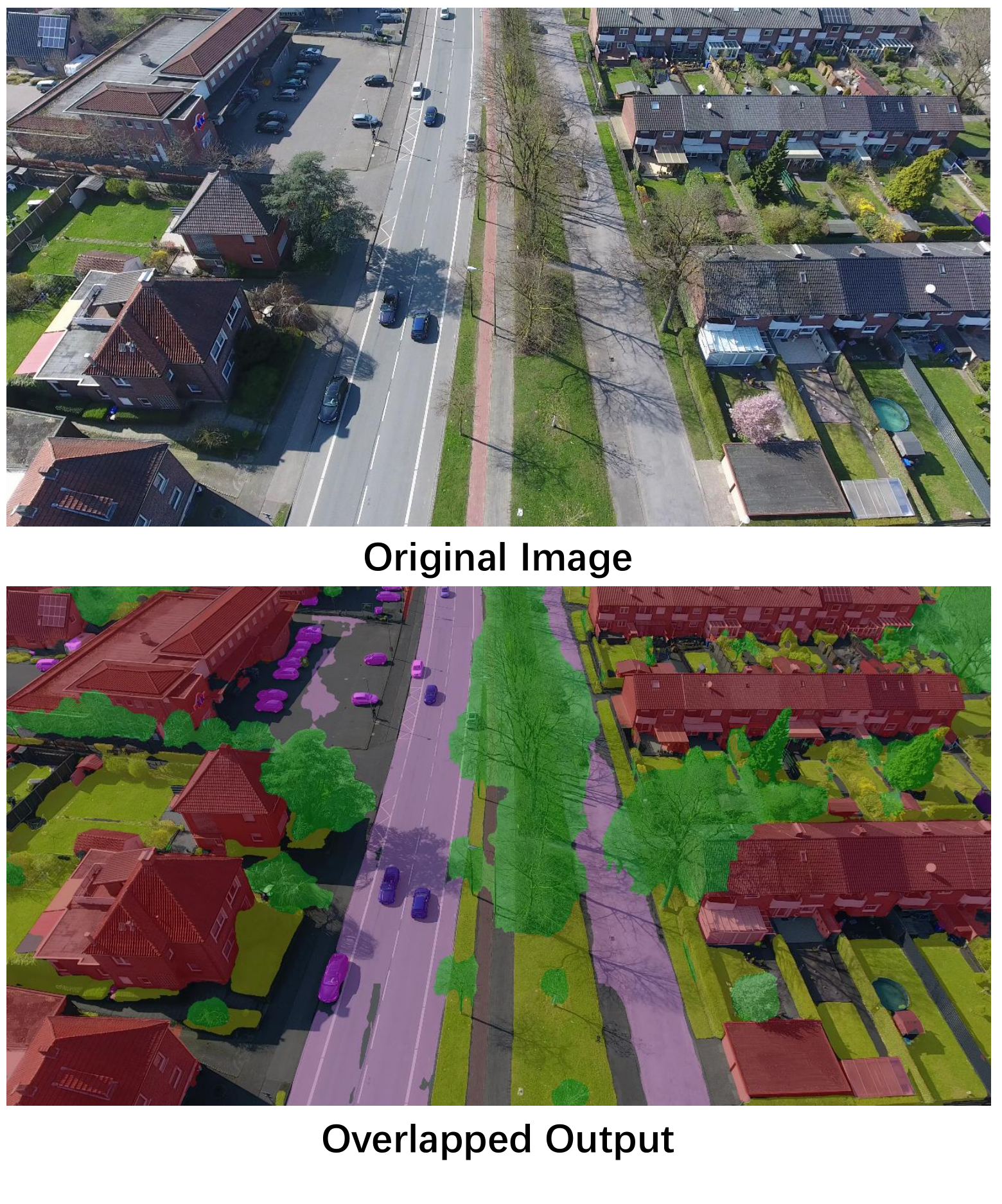}
	\end{minipage}
	\begin{minipage}{.66\linewidth}
		\includegraphics[width=1.0\linewidth]
		{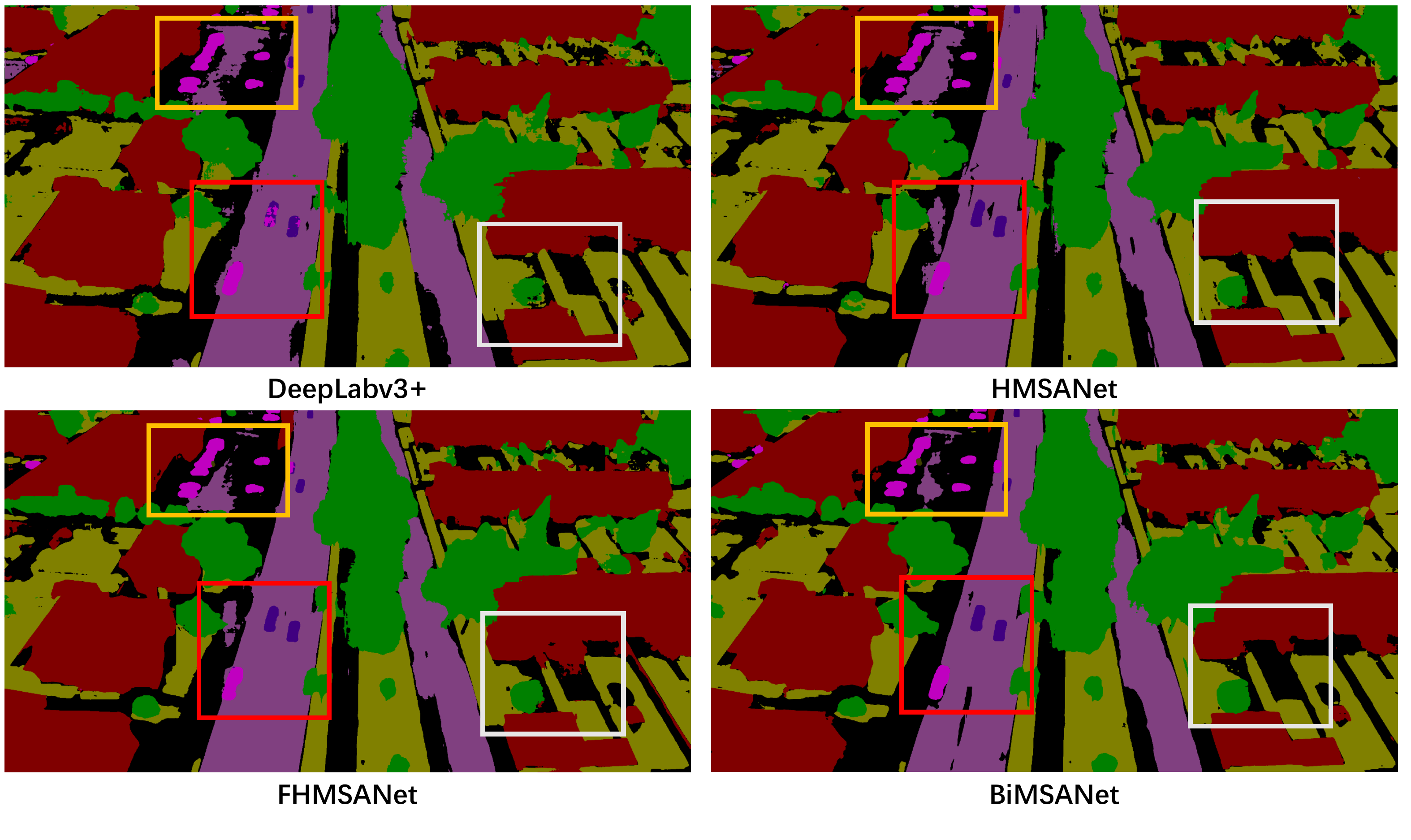}
	\end{minipage}
	\vspace{-3mm}
	\caption{Qualitative comparisons of different models on the UAVid2020 test set. The example image is from the test set (seq30, 000400). Bottom left image shows the overlapped result of the BiMSANet output and the original image. Three example regions for comparisons are marked in red, orange, and white boxes.}
	\label{fig:cmp}
\end{figure*}

\subsection{Implementation}
\textbf{Training.}
All the models in the experiments are implemented with pytorch~\cite{pytorch}, and trained on a single Tesla V100 GPU of $16G$ memory with a batch of $2$ images. Mixed precision and synchronous batch normalization are applied for the model. Stochastic gradient decent with a momentum $0.9$ and weight decay of $5e^{-4}$ is applied as the optimizer for training. "Polynomial" learning rate policy is adopted~\cite{poly} with a poly exponent of $2.0$. The initial learning rate is set to $5e{-3}$. The model is trained for $175$ epochs by random image selection. We apply random scaling for the images from $0.5\times$ to $2.0\times$. Random cropping is applied to acquire image patches of size of $896\times896$.

\textbf{Testing.}
As the $4K$ image is too large to fit into the GPU, we apply cropping during testing as well. The image is partitioned into overlapped patches for evaluation as in ~\cite{uavid} and the average of the score maps are used for the final output in the overlapped regions. The crop size is set to $896\times896$ with an overlap of $512$ pixels in both horizontal and vertical directions.

\subsection{Model Comparisons}
In this section, we will presents the semantic segmentation results on the test set of UAVid2020 dataset for multi-scale-dilation net (MSDNet)~\cite{uavid}, deeplabv3+~\cite{deeplabv3+}, hierarchical multi-scale attention net (HMSANet)~\cite{hmsa}, feature level hierarchical multi-scale attention net (FHMSANet), and our proposed bidirectional multi-scale attention networks (BiMSANet). MSDNet is included as reference, which uses an old trunk FCN-8s~\cite{fcn8s} in each scale. The major comparisons are among DeepLabv3+, HMSANet, FHMSANet, and BiMSANet.

The mIoU scores and the IoU scores for each individual class are shown in Table~\ref{tab:comparisons}. Among all the compared models, the BiMSANet performs the best regarding the mIoU metric. Our BiMSANet has a more balanced prediction ability for both large and small objects. 

For the evaluation of each individual class, the BiMSANet ranks the first for classes of clutter, building, tree, static car, and human. The most distinct improvement is for the static car, which is $2.72\%$ higher than the second best score. With only the context information, our method could achieve decent scores for classes of both moving car and static car.

For human class, the scores of HMSANet, FHMSANet and BiMSANet are all significantly higher than the DeepLabv3+, which shows the superiority of multi-scale attention mechanism in handling the small objects. Thanks to the bidirectional multi-scale attention design, BiMSANet achieves the best performance for the human class.

Qualitative comparisons are shown in Figure~\ref{fig:cmp}. The example image is selected from the test set (seq30, 000400). As the ground truth label is reserved for benchmark evaluation, the overlapped output is shown instead in Figure~\ref{fig:cmp}. Three example regions are marked in red, orange, and white boxes. 

In the red box region, it could be seen that the deeplabv3+ struggles to give coherent predictions for cars in the middle of the road, while the other three models have better results due to the multi-scale attention. The HMSANet and the FHMSANet wrongly classify part of the sidewalks, which is outside the road, as road class. BiMSANet handles better in this area. However, part of the road near the lane-mark are wrongly classified as clutter by the BiMSANet.
In the orange box region, the parking lot, which belongs to the clutter class, is predicted as the road by all four models, and the BiMSANet makes the least error.
In the white box region, the ground in front of the entrance door is wrongly classified as building by all models except the BiMSANet. This is benefited from the bidirectional multi-scale attention design.

\begin{figure*}[ht!]
	\centering
	\begin{minipage}{.33\linewidth}
		\includegraphics[width=1.0\linewidth]
		{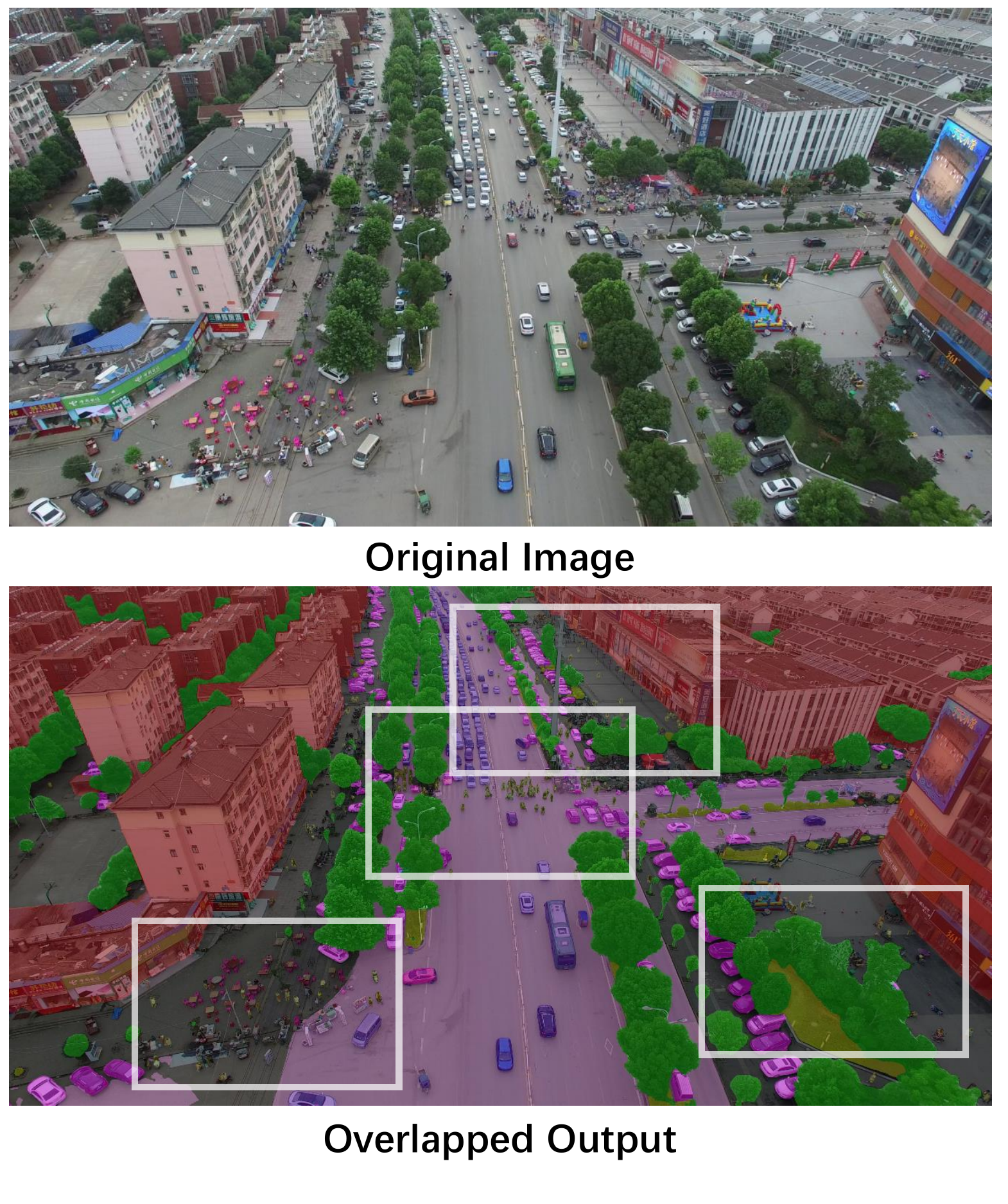}
	\end{minipage}
	\begin{minipage}{.65\linewidth}
		\includegraphics[width=1.0\linewidth]
		{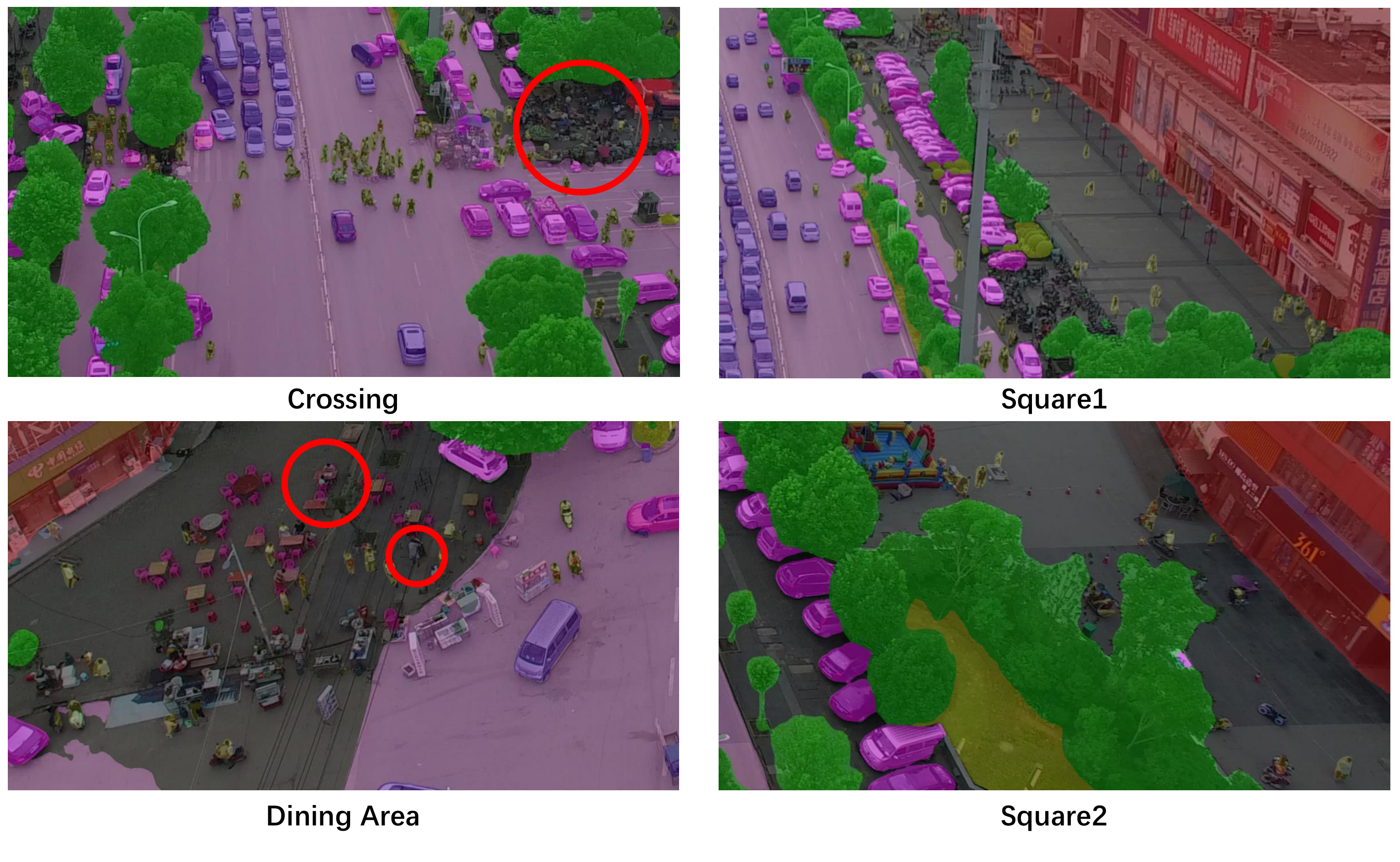}
	\end{minipage}
	\vspace{-2mm}
	\caption{Qualitative example of human class segmentation by the BiMSANet. The example image is from the test set (seq22, 000900). The left column shows the original full image and the overlapped output. The middle and the right columns show the image patches cropped from the overlapped output (marked by white boxes), which all focus on the human class. The red circles mark some missing segmentation.}
	\label{fig:human}
\end{figure*}

We have also shown the performance for human class segmentation in Figure~\ref{fig:human}. The example image is from the test set (seq22, 000900). The zoomed in images in the middle and the right columns correspond to the patches in the white boxes of the overlapped output. The four patches are from different context, which is very complex in some local regions. Even though the humans in the image are quite small and in many different poses, such as standing, sitting, and riding, our model can still effectively detect and segment most of the humans in the image.

\begin{table*}[h]
	\centering
	\begin{tabular}{|l|c|c|c|c|c|c|}\hline
		Methods&mIoU(\%)&mIoU Gains(\%)&Trunk&Multi-Scale Attention&Feature Level Fusion&Bidirection\\\hline
		DeepLabv3+&67.36&-&\cmark&-&-&-\\
		HMSANet&70.03&+2.67&\cmark&\cmark&-&-\\
		FHMSANet&70.33&+0.30&\cmark&\cmark&\cmark&-\\
		BiMSANet&70.80&+0.47&\cmark&\cmark&\cmark&\cmark\\\hline
	\end{tabular}
	\caption{Ablation study for models. The performance gains could be observed by gradually adding components.}
	\label{tab:ablation}
\end{table*}

\subsection{Ablation Study}
In this section, we will compare the performance gains by gradually adding the components. The corresponding results are shown in Table~\ref{tab:ablation}. It is easy to see that the multi-scale processing is useful for the oblique view UAV images. The mIoU score has increased by $2.67\%$ by including the multi-scale attention into the networks. The feature level fusion is also proved to be useful as it helps the networks to improve the mIoU score by $0.3\%$. By further adding the bidirectional attention mechanism, the networks improve the mIoU score by another $0.47\%$.

\subsection{Analysis of Learned Multi-Scale Attentions}
In this section, we will analyze the learned multi-scale attentions from the BiMSANet to better understand how the attentions work. We explore from mainly three perspectives: attentions of different channels, different scales, and different directions. The example image is from the test set (seq25,000400). Attentions from both Attn1 branch and Attn2 branch are used, noted as $\alpha$ and $\beta$ in Figure~\ref{fig:bimsa}. $\alpha$ is for the fine to coarse pathway, while $\beta$ is for the coarse to fine pathway.

\begin{figure*}[ht!]
	\centering
	\includegraphics[width=.85\linewidth]{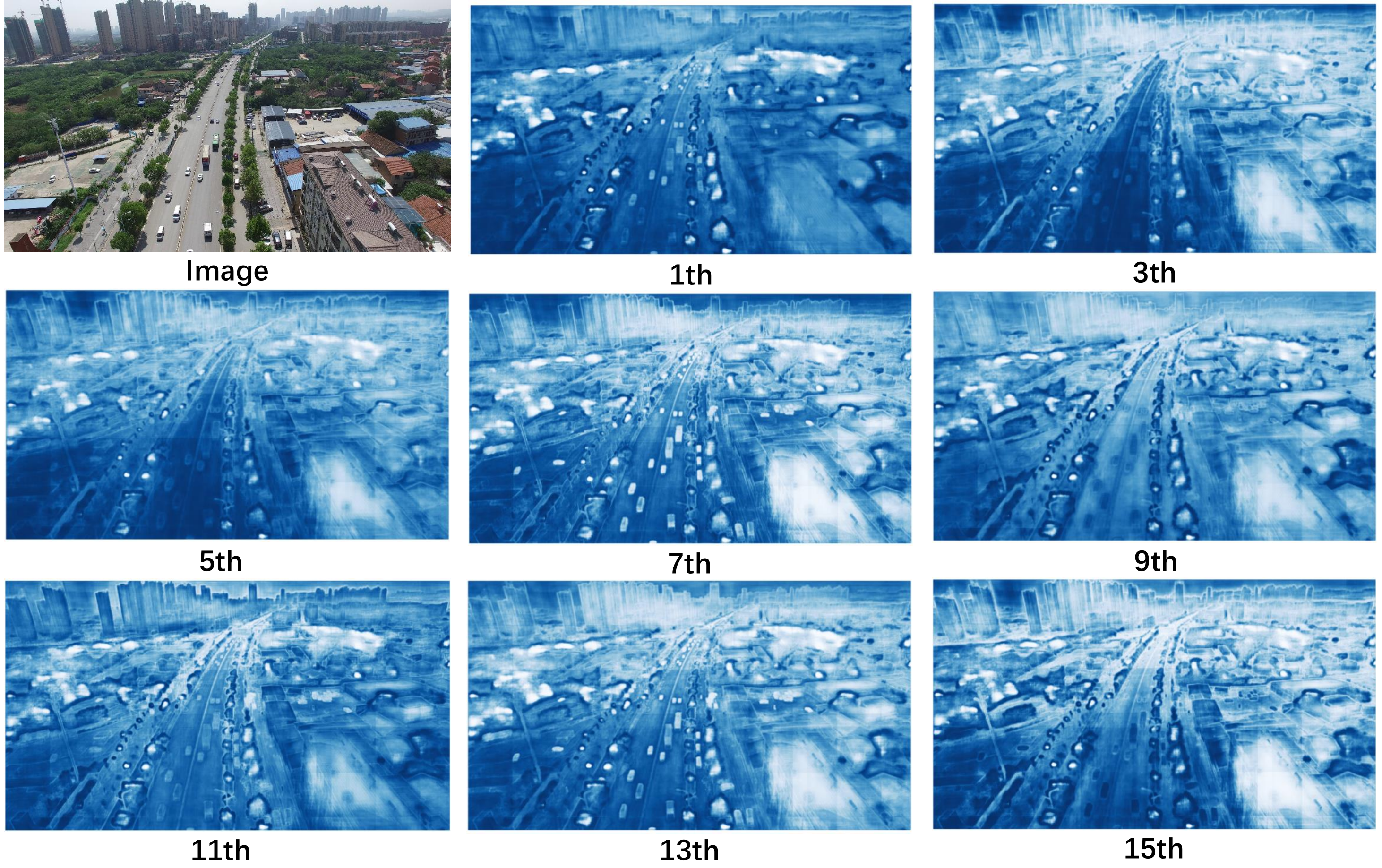}
	\caption{Attention analysis of different channels. Example attentions are of $1\times$ scale from Attn1 branch. The image on the top left shows the image adopted. The other $8$ images are the attention maps from different channels. Channel indices are presented below the images. Brighter color means higher value. Best visualized with zoom in.}
	\label{fig:attn_ch}
\end{figure*}
\subsubsection{Attention of different channels}
The multi-scale attentions in our BiMSANet have $\frac{1}{2}nc$ channels ($16$ in our case), which is different from the HMSANet~\cite{hmsa}, whose attention has only one single channel for all classes. The attentions guide the fusion of features across scales. Example attentions of different channels in $1\times$ scale branch are shown in Figure~\ref{fig:attn_ch}. Different channels have different attentions focusing on different parts of the image. It is obvious that different channels have different focus for different classes, e.g., $1th$ channel more focus on trees, $3th$ channel less focus on roads, and $7th$ channel have the most focus on moving cars.

\begin{figure*}[ht!]
	\centering
	\begin{minipage}{.49\linewidth}
		\includegraphics[width=1.0\linewidth]
		{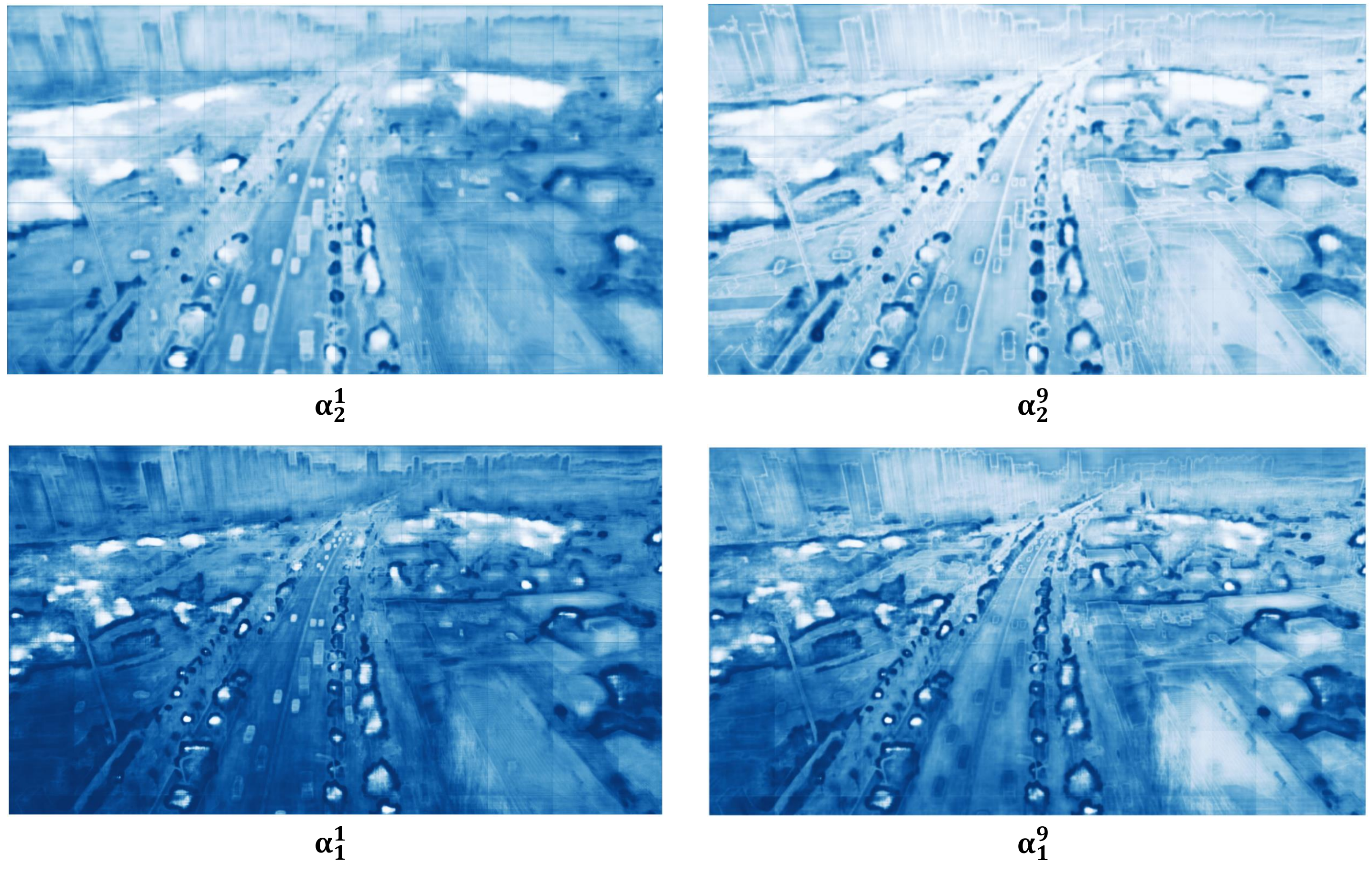}
	\end{minipage}
	\begin{minipage}{.49\linewidth}
		\includegraphics[width=1.0\linewidth]
		{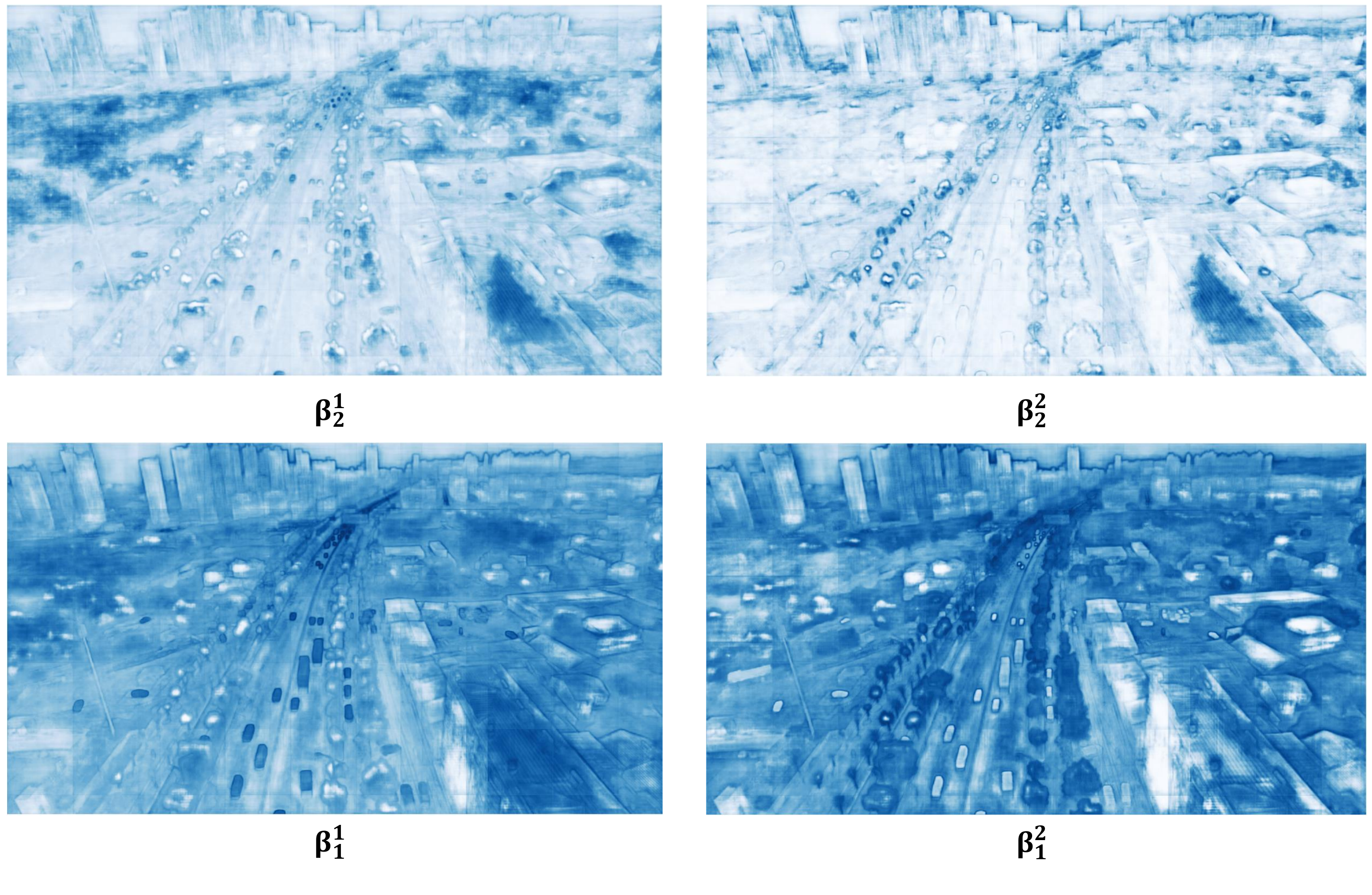}
	\end{minipage}
	\vspace{-2mm}
	\caption{Attention analysis of different scales. We select $4$ attentions from each of the Attn1 branch and the Attn2 branch. $\alpha, \beta$ are of the same meaning as in Figure~\ref{fig:bimsa}. The superscripts are the channel index of the attentions. $\alpha_{2}, \beta_{2}$ correspond to the attentions predicted in the $0.5\times$ scale and the $2\times$ scale. $\alpha_{1}, \beta_{1}$ are predicted in $1\times$ scale. Brighter color means higher value. Best visualized with zoom in.}
	\label{fig:attn_sc}
\end{figure*}
\begin{figure*}[ht!]
	\centering
	\includegraphics[width=1.0\linewidth]{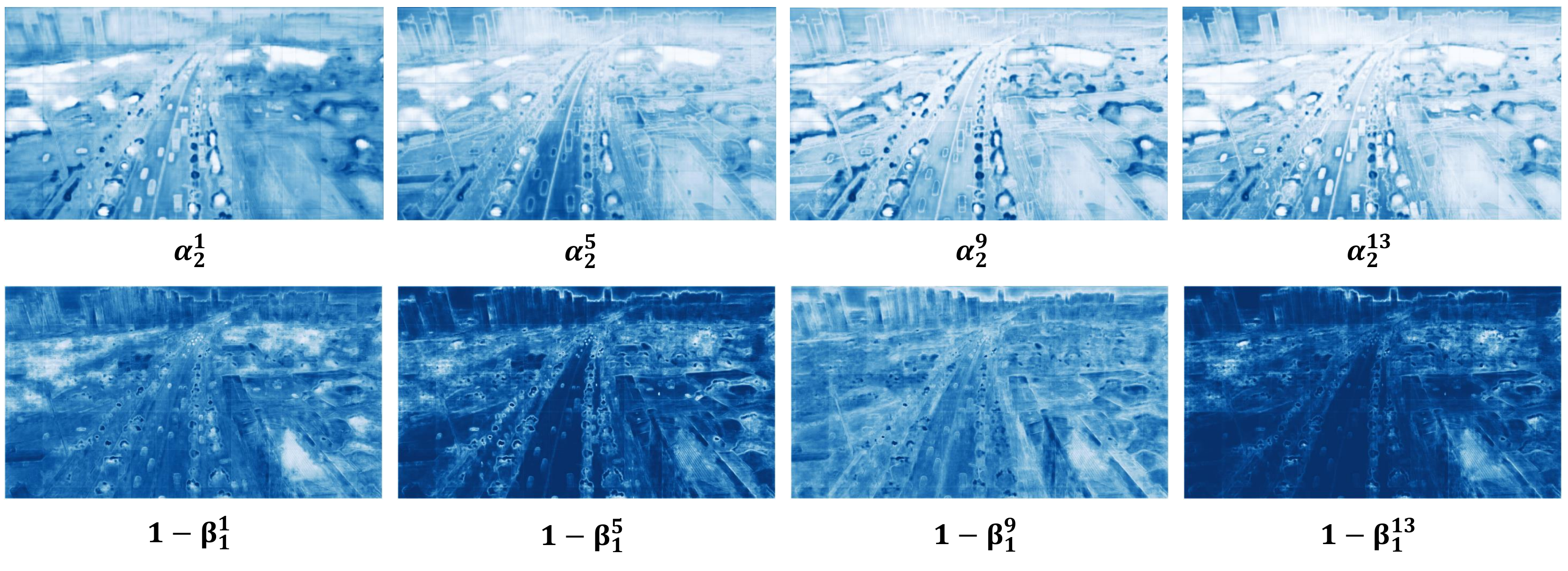}
	\caption{Attention analysis of different directions. The figure shows the attentions for fusing features of scale $0.5\times$ and scale $1\times$. $\alpha_{2}$ is for the fine to coarse pathway, while $1-\beta_{1}$ is for the coarse to fine pathway. Brighter color means higher value. Best visualized with zoom in.}
	\label{fig:attn_bi}
\end{figure*}
\subsubsection{Attention of different scales}
In order to analyze the difference of attentions in different scales, we have selected $4$ attentions from each of the Attn1 branch and the Attn2 branch as shown in Figure~\ref{fig:attn_sc}. The superscripts are the channel index of the attentions. By comparing the $\alpha_{1}$ with $\alpha_{2}$, which are predicted in $1\times$ and $0.5\times$ scales, we could see that attentions in different scales have different focus. The difference of the same channel between $\alpha_{1}$ and $\alpha_{2}$ are more worth of comparisons. The same applies for $\beta_{1}$ and $\beta_{2}$.

From $\alpha^{1}_{1}$ and $\alpha^{1}_{2}$, it could be noted that the recognition of cars in closer distance are more based on context, since the values of $\alpha^{1}_{2}$ are larger than $\alpha^{1}_{1}$. The recognition of road that are closer to the camera also relies more on the coarser level features, which is reasonable as the road area is large and requires more context for recognition. It is also interesting to note that the middle lane-marks is even brighter than other parts of the road in $\alpha^{1}_{2}$, which means the recognition requires more context. It is reasonable as the color and the texture of the lane-marks are quite different compared to other parts of the road. The distant buildings near the horizon relies more on the coarser level features as well.

We have also noticed that the $\alpha_{2}$ ($0.5\times$ scale) and $\beta_{2}$ ($2\times$ scale) have larger values on average compared with $\alpha_{1}$ and $\beta_{1}$ ($1\times$ scale), which means that features with context information and fine details are both valuable for object recognition.

\subsubsection{Attention of different directions}
In our bidirectional design, both the coarse to fine pathway and the fine to coarse pathway fuse the features from three scales ($0.5\times,1\times,2\times$). In this section, we analyze if the feature fusion in two pathways has the same attention pattern. Attention examples are shown in Figure~\ref{fig:attn_bi}. Attentions $\alpha_{2}$ and $1-\beta_{1}$ from two pathways are both for the feature fusion across scale $0.5\times$ and $1\times$. Although the attention values of same pixels can not be directly compared as the feature sources are different (Feat1 and Feat2), it is still evident that the attention densities on average are quite different. There are more activation in $\alpha_{2}$ than $1-\beta_{1}$, showing that the two pathways play different roles for feature fusion across same scales.

\section{Conclusion}
In this paper, we have proposed the bidirectional multi-scale attention networks (BiMSANet) for the semantic segmentation task. The hierarchical design adopted from ~\cite{hmsa} allows the usage of larger trunk for better performance. The feature level fusion and the bidirectional design allows the model to more effectively fuse the features from both of the adjacent coarser scale and the finer scale. We have conducted the experiments on the UAVid2020 dataset~\cite{uavid}, which have large variation in spatial resolution. The comparisons among different models have shown that our BiMSANet achieves better results by balancing the performance of small objects and large objects. Our BiMSANet achieves the state-of-art result with a mIoU score of $70.80\%$ for the UAVid2020 benchmark.

{
	\begin{spacing}{1.17}
		\normalsize
		\bibliography{ISPRSguidelines_authors} 
	\end{spacing}
}

\end{document}